\pdfoutput=1

\documentclass[11pt]{article}

\usepackage[preprint]{acl}

\usepackage{times}
\usepackage{latexsym}

\usepackage[T1]{fontenc}

\usepackage[utf8]{inputenc}

\usepackage{microtype}

\usepackage{inconsolata}
\usepackage{comment}

\usepackage{graphicx}
\usepackage{multirow}
\usepackage{xcolor}
\usepackage{subcaption}
\usepackage{booktabs}
\usepackage{amsmath}

\usepackage{float}

\usepackage{makecell}
\newcommand{\fromdemo}[1]{\textcolor{violet}{#1}}  
\usepackage[greek,english]{babel}

\title{Wikipedia is Not a Dictionary, Delete! Text Classification as a Proxy for Analysing Wiki Deletion Discussions}

\author{
\textbf{Hsuvas Borkakoty\textsuperscript{1}} and
\textbf{Luis Espinosa-Anke\textsuperscript{1,2}}
\\
\textsuperscript{1}Cardiff NLP, School of Computer Science and Informatics, Cardiff University, UK 
\\
\textsuperscript{2}AMPLYFI, UK
\\
\small{
   \texttt{\{borkakotyh,espinosa-ankel\}@cardiff.ac.uk}
 }
}

\begin{document}

\maketitle
\begin{abstract}

Automated content moderation for collaborative knowledge hubs like Wikipedia or Wikidata is an important yet challenging task due to multiple factors
. In this paper, we construct a database of discussions happening around \textit{articles marked for deletion} in several Wikis and in three languages, which we then use to evaluate a range of LMs on different tasks (from predicting the outcome of the discussion to identifying the implicit policy an individual comment might be pointing to). Our results reveal, among others, that discussions leading to deletion are easier to predict, and that, surprisingly, self-produced tags (keep, delete or redirect) don't always help guiding the classifiers, presumably because of users' hesitation or deliberation within comments\footnote{Dataset available at: \url{https://huggingface.co/datasets/hsuvaskakoty/wider}.}. 
\end{abstract}


\section{Introduction}
Wikipedia and its sister Wikis play an indispensable role as a collaborative knowledge source, and are widely used by students \cite{selwyn2016students} and the general public  \cite{singer2017we, lemmerich2019world} alike. They fulfill use cases that range from core knowledge \textit{go-tos}, as well as ``free'' supporting documentation for content providers and search engines (e.g., Google\footnote{\url{https://en.wikipedia.org/wiki/Relationship\_between\_Google\_and\_Wikipedia}} or YouTube\footnote{\url{https://support.google.com/youtube/answer/7630512?hl=en}}). However, due to their size and, most importantly, their collaborative nature, ensuring high quality in these platforms is challenging, especially given the need to ``map'' content to existing policies at least semi-automatically \cite{ribeiro2022automated}. This is particularly relevant in the GenAI era, as AI-generated content has proliferated throughout the Internet \cite{brooks2024rise}.

\begin{figure}[!t]
    \centering
    \includegraphics[width=\columnwidth]{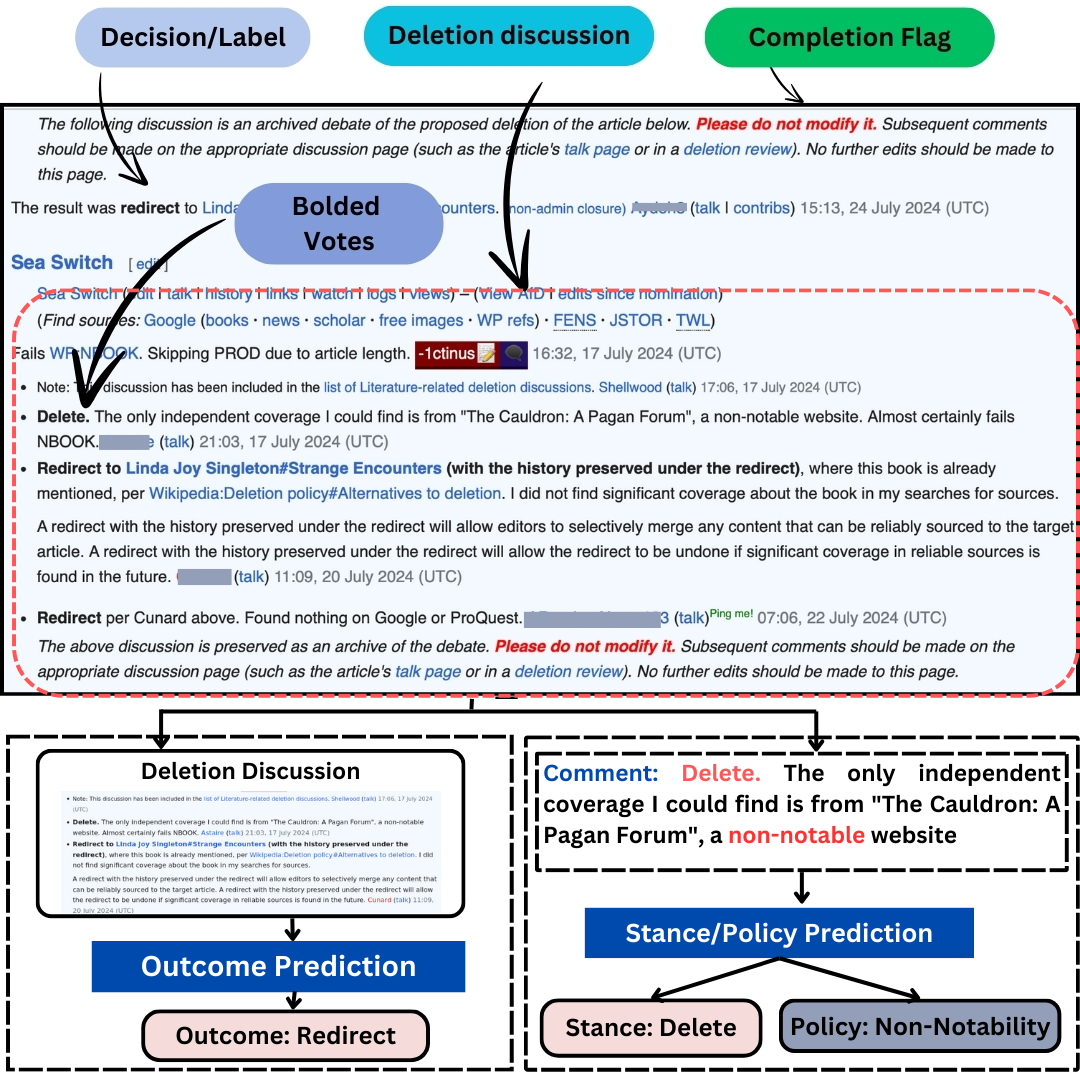}
    \caption{Example of a Deletion Discussion in English Wikipedia}
    \label{fig:deletion_discussion_example}
\end{figure}    

\begin{table*}[!t]
\resizebox{\textwidth}{!}{%
\begin{tabular}{p{2cm} p{1cm} p{2.2cm} p{11cm}} 
\hline
\textbf{Task}                       & \textbf{Lang.}      & \textbf{Platform}   & \textbf{Label Set}                                                                 \\ \hline
\multirow{7}{*}{\makecell{Outcome \\ Prediction} } & \multirow{5}{*}{En} & Wikipedia           & delete, keep, redirect, no-consensus, merge, speedy keep, speedy delete, withdrawn \\ \cline{3-4} 
                                    &                     & Wikidata-ent.       & delete, keep, merge, redirect, no-consensus, comment                               \\ \cline{3-4} 
                                    &                     & Wikidata-pr.        & delete, keep, no-consensus                                                         \\ \cline{3-4} 
                                    &                     & Wikiquote           & delete, keep, redirect, merge, no-consensus                                        \\ \cline{3-4} 
                                    &                     & Wikinews            & delete, keep, speedy delete, comment                                               \\ \cline{2-4} 
                                    & Es                  & Wikipedia           & borrar (delete), mantener (keep), fusionar (merge), otros (others)                 \\ \cline{2-4} 
                                    & Gr                  & Wikipedia           & \textgreek{Διαγραφή} (delete), \textgreek{Διατήρηση} (keep), \textgreek{Δεν υπάρχει συναίνεση} (no-consensus)                                                         \\ \hline
Stance Detection                    & En                  & Wikipedia           & delete, keep, merge, comment                                                       \\ \hline
Policy Prediction                    & En                  & Wikipedia           &  Wikipedia:Notability, Wikipedia:What Wikipedia is not, Wikipedia:No original research, Wikipedia:Verifiability, Wikipedia:Arguments to avoid in deletion discussions, Wikipedia:Biographies of living persons, Wikipedia:Criteria for speedy deletion, Wikipedia:Articles for deletion, Wikipedia:Wikipedia is not a dictionary, Wikipedia:Deletion policy   \\ \hline
\end{tabular}%
}
\caption{Label set for the different tasks and datasets we consider in this paper (Outcome, Stance, and Policy), for three languages (English: En, Spanish: Es, and Greek: Gr) and five (4+1) platforms (Wikipedia, Wikidata-Entity (ent.) and Property (pr.), Wikinews, and Wikiquote).}
\label{tab:labels}
\end{table*}

More generally, content moderation in online platforms is often the outcome of group coordination and communication \cite{chidambaram2005out,jensen2005collaboration,butler2008don}. Unsurprisingly, NLP plays an important role in automating this process. For example, \citet{singhal2023sok} defined a framework for social media moderation as a function of community guidelines, policy enforcement and violation detection. Some prominent examples of works connected with such a framework are: policy based content moderation in Facebook \cite{sablosky2021dangerous}, rule-breaking behavior analysis on Reddit \cite{chandrasekharan2018the}, and topic based content moderation discourse on X \cite{alizadeh2022content}. In the case of Wikis, both the guidelines and the rules that govern the quality of their content are maintained by contributions from the community \cite{seering2020reconsidering}, where discussion-based approaches towards content moderation are the norm. The way this generally works is that, given an article flagged by a community member, users justify their stance towards it and its adherence to the policies, and then editors act. However, manual efforts to clear the backlog are insufficient. Therefore, NLP techniques for predicting the outcome of a deletion discussion, or for capturing a user's stance towards a specific article are critical \cite{mayfield2019stance,kaffee2023should}. 
 Despite this need, there is a surprising lack of work beyond Wikipedia, and there is almost no published work that looks at non-English languages (with the exception of \citet{kaffee2023should}). Moreover, in-depth comparative analyses of parameter-efficient techniques have also been so far largely unexplored.



We therefore aim to address all of the above with an analysis on a novel collection of deletion discussions, in three languages and four platforms. These discussions come, if resolved, alongside the discussion outcomes (generally speaking, suggesting to \texttt{keep} or \texttt{delete} the article, although the actual outcome tags are more fine grained than this), with individual comments having their own stance and referring to specific policies \citet{kaffee2023should}. Our classification experiments set strong baseline results for the community to build upon, and provide insights into these community-led activities.

\begin{table}[!t]
\centering
\scriptsize 
\resizebox{\columnwidth}{!}{%
\begin{tabular}{llr}
\hline
\textbf{Language} & \textbf{Platform}           & \textbf{Total}    \\ \hline
\multirow{5}{*}{en} 
 & Wikipedia           & 18,528  \\ 
 & Wikidata-entities   & 355,428 \\ 
 & Wikidata-properties & 498    \\ 
 & Wikinews            & 91     \\ 
 & Wikiquote           & 695    \\ \hline
es & Wikipedia           & 3,274   \\ \hline
gr & Wikipedia           & 392    \\ \hline
\end{tabular}%
}
\caption{Overall number of deletion discussions per language and platform in the outcome prediction dataset.}
\label{tab:overall_data_stats}
\end{table}

\begin{table*}[!t]
\centering
\renewcommand{\arraystretch}{1.3} 
\small 
\resizebox{\textwidth}{!}{%
\begin{tabular}{@{}l p{1.8cm} p{3.5cm} p{6.5cm} p{2.5cm} @{}}
\toprule
\multicolumn{1}{c}{\textbf{Platform}} &
  \textbf{Language} &
  \multicolumn{1}{c}{\textbf{Example title}} &
  \multicolumn{1}{c}{\textbf{Discussion (truncated)}} &
  \multicolumn{1}{c}{\textbf{Outcome}} \\ \midrule
Wikipedia &
  en &
  Beast Poetry &
  \begin{tabular}[c]{@{}p{6.5cm}@{}}\raggedright Editor 1: Keep in one form or another.\\ Editor 2: One option could be to re-frame the article to be about the book.\end{tabular} &
  Keep \\\hline
Wikidata-Ent &
  en &
  Q28090948 &
  \begin{tabular}[c]{@{}p{6.5cm}@{}}\raggedright Editor 1: no description : Vandalism.\end{tabular} &
  Delete \\\hline
Wikidata-Prop &
  en &
  JMdict sequence number (P11700) &
  \begin{tabular}[c]{@{}p{6.5cm}@{}}\raggedright Editor 1: Deleted - ( ) Support I assume no comments have been made because this is a clear case to delete...\end{tabular} &
  Delete \\\hline
Wikinews &
  en &
  Mugalkhod Jeedga Mutta organizes mass marriage in Belgaum, India &
  \begin{tabular}[c]{@{}p{6.5cm}@{}}\raggedright Editor 1: There is no further meaningful work on the article.\\ Editor 2: All advice ignored, kill it with cleansing fire and stop wasting time.\end{tabular} &
  Speedy delete \\\hline
Wikiquote &
  en &
  3rd Rock From The Sun &
  \begin{tabular}[c]{@{}p{6.5cm}@{}}\raggedright Editor 1: Two reasons to delete this:\\ - It is a copy of about half of quotes on IMDB\\ - 3rd Rock from the Sun is a different article.\\ Editor 2: Merge, perhaps with some trimming.\end{tabular} &
  Merge \\\hline
Wikipedia-Es &
  es &
  Héroes: El legado de la Evolución &
  \begin{tabular}[c]{@{}p{6.5cm}@{}}\raggedright Editor 1: Bórrese Irrelevante enciclopédico.\\ Editor 2: Bórrese Irrelevante.\end{tabular} &
  Borrar \\\hline
Wikipedia-gr &
  gr &
  \textgreek{Μουσείο Μιχάλη Τσαρτσίδη} &
  \begin{tabular}[c]{@{}p{6.5cm}@{}}\raggedright Editor 1: \textgreek{Σχόλιο Διαφωνώ έντονα με την λογική/φράση «Απλώς ένα από τα πολλά ανά την Ελλάδα μουσε}...\\ Editor 2: \textgreek{Ο μόνος λόγος διαγραφής μπορεί να είναι η παραβίαση πνευματικών}.\end{tabular} &
  \textgreek{διαγραφή} \\\bottomrule
\end{tabular}%
}
\caption{Examples for different platforms and languages, alongside outcome labels.
}
\label{tab:dataset_examples_all}
\end{table*}

\section{Tasks and dataset Construction} 
\label{sec:dataset}

We consider three tasks, namely (1) \textbf{Outcome prediction} - given a full discussion around an article marked for deletion, predict the final decision; (2) \textbf{Stance detection}, i.e. given an individual comment, determine its stance towards the decision to be made for that article; and (3) \textbf{Policy prediction}, where again, given one single comment, we want to determine the policy that comment is most likely be referring to (Figure \ref{fig:deletion_discussion_example} shows an exmaple). We build a novel dataset for outcome prediction, while for the other two tasks we largely rely on the dataset from \citet{kaffee2023should} (although with some important modifications to enable the goal of this paper, namely an in-depth analysis). We provide more detail about these datasets in the following subsections.

\subsection{Outcome Prediction}
We retrieve and clean deletion discussions programmatically\footnote{We use the \textsc{wide-analysis} toolkit: https://pypi.org/project/wide-analysis/ \cite{borkakoty2024wide}.} for three different languages and four platforms (with Wikidata being split in two: properties and entities, as their discussions happen separately). In terms of coverage, for small platforms with less activity such as Wikinews or Wikiquote\footnote{According to Wikimedia Statistics, in 2024 Wikipedia and Wikidata received 130 Billion and 3 Billion pageviews, whereas Wikinews and Wikiquote received 77 Million and 179 Million respectively.}, we consider all data available in their website at the time of scraping, whereas for larger and more active platforms like Wikidata, we consider the last 4 full years (from 2021 to 2024, both inclusive). Label sets per task are provided in Table \ref{tab:labels}, whereas statistics in terms of raw size can be found in Table \ref{tab:overall_data_stats}.

\subsection{Dataset for Stance Detection and Policy Prediction}
\label{subsec:stance_policy_data}

For stance detection and policy prediction we use the existing \textsc{wiki-stance} dataset \cite{kaffee2023should}. We keep the stance detection dataset as-is, including their original label set. However, for policy prediction, we consider a reduced label set in order to perform error analysis, and so keep only the top 10 most frequent labels (as opposed to the 92 contained in the original dataset). This change, however, has a small effect on the overall dataset size, as retaining only these labels still results in roughly 80\% of the original dataset. As an example, policy labels contained in the original dataset like \texttt{Wikipedia: Userfication}, \texttt{Wikipedia: Record charts} or \texttt{Wikipedia: Attack page} account for only 106, 105 and 102 instances, respectively (out of 437,770, which means a negligible percentage, around 0.02\%).

\section{Experiments}

With these datasets in place, we proceed to run classification experiments.

\subsection{Outcome Prediction}
\label{sec:outcome_prediction}

\begin{table*}[!t]
\centering
\scriptsize
\resizebox{\textwidth}{!}{
\begin{tabular}{@{}llrrrrr@{}}
\toprule
\multicolumn{1}{c}{\textbf{Input Type}} &
  \multicolumn{1}{c}{\textbf{Model}} &
  \textbf{Wikip.} &
  \textbf{Wikid.-ent} &
  \textbf{Wikid.-pr} &
  \textbf{Wikin.} &
  \textbf{Wikiq.} \\ \midrule
\multirow{6}{*}{Fulltext} & RoBERTa-B     & 0.56          & 0.61          & 0.56         & 0.4           & 0.71          \\
 &
  RoBERTa-L &
  \textbf{0.58} &
  \textbf{0.63} &
  \textbf{0.62} &
  \textbf{0.47} &
  \textbf{0.76} \\
                          & BERT-B        & 0.56          & 0.57          & 0.54         & 0.4           & 0.7           \\
                          & BERT-L        & \textbf{0.58} & 0.62          & 0.61         & \textbf{0.47} & 0.73          \\
                          & DistilBERT    & 0.55          & 0.56          & 0.5          & 0.39          & 0.57          \\
                          & Tw.-RoBERTa-B & 0.49          & 0.6           & 0.55         & 0.4           & 0.7           \\ \midrule
\multirow{6}{*}{Masked}   & RoBERTa-B     & 0.49          & 0.51          & 0.56         & 0.36          & 0.62          \\
                          & RoBERTa-L     & \textbf{0.52} & \textbf{0.61} & 0.56         & \textbf{0.42} & 0.65          \\
                          & BERT-B        & 0.49          & 0.57          & 0.56         & 0.33          & 0.7           \\
                          & BERT-L        & 0.5           & 0.62          & \textbf{0.6} & 0.36          & \textbf{0.72} \\
                          & DistilBERT    & 0.43          & 0.56          & 0.5          & 0.27          & 0.32          \\
                          & Tw-RoBERTa-B  & 0.46          & 0.52          & 0.42         & 0.3           & 0.38          \\ \bottomrule
\end{tabular}
}
\caption{F1 Scores for fine-tuned models in Wikipedia (Wikip.), Wikid.-end (Wikidata, entities subset), Wikid.-pr (Wikidata, entities subset), and Wikiq. (Wikiquote), both for (a) full text and (b) masked text inputs. Models are identified by their versions: Tw (Twitter), B (Base) and L (Large).}
\label{tab:res_combined}
\end{table*}

While previous works \cite{mayfield2019analyzing} cast outcome prediction as binary classification (\texttt{Delete} and \texttt{Keep}), we follow Wikipedia's official guidelines\footnote{\url{https://en.wikipedia.org/wiki/Wikipedia:Guide_to_deletion}} and propose a more nuanced scheme (again, c.f. Table \ref{tab:labels}) and re-cast it as a multi-class classification. Following \citet{mayfield2019stance}, we have two set ups: \textit{Masked}, where labels are redacted from the comments, and \textit{Fulltext} where classifiers see the full dataset, including the self-assigned labels (which in theory act as extremely informative features about the stance of each comment and therefore good predictors of the final outcome of the discussion - however, as we will see, this is not always the case). We evaluate BERT \cite{devlin2018bert}, RoBERTa \cite{liu2019roberta}, DistilBERT \cite{sanh2020distilbert} (all in Base and Large), and Twitter-RoBERTa-Base \cite{barbieri2020tweeteval} (in order to explore the effect of models tailored to user-generated content). Information about training, validation and tests splits, and implementation details, are provided in Appendices \ref{sec:app-data} (Table \ref{tab:data_stats_splits}) and \ref{app:finetuning_lm}, respectively.  Furthermore, we divide our set of experiments in three scenarios: in-platform, cross-platform and multilingual.

\begin{figure}[!h]
    \centering
    \begin{tabular}{c}
        \includegraphics[width=\columnwidth]{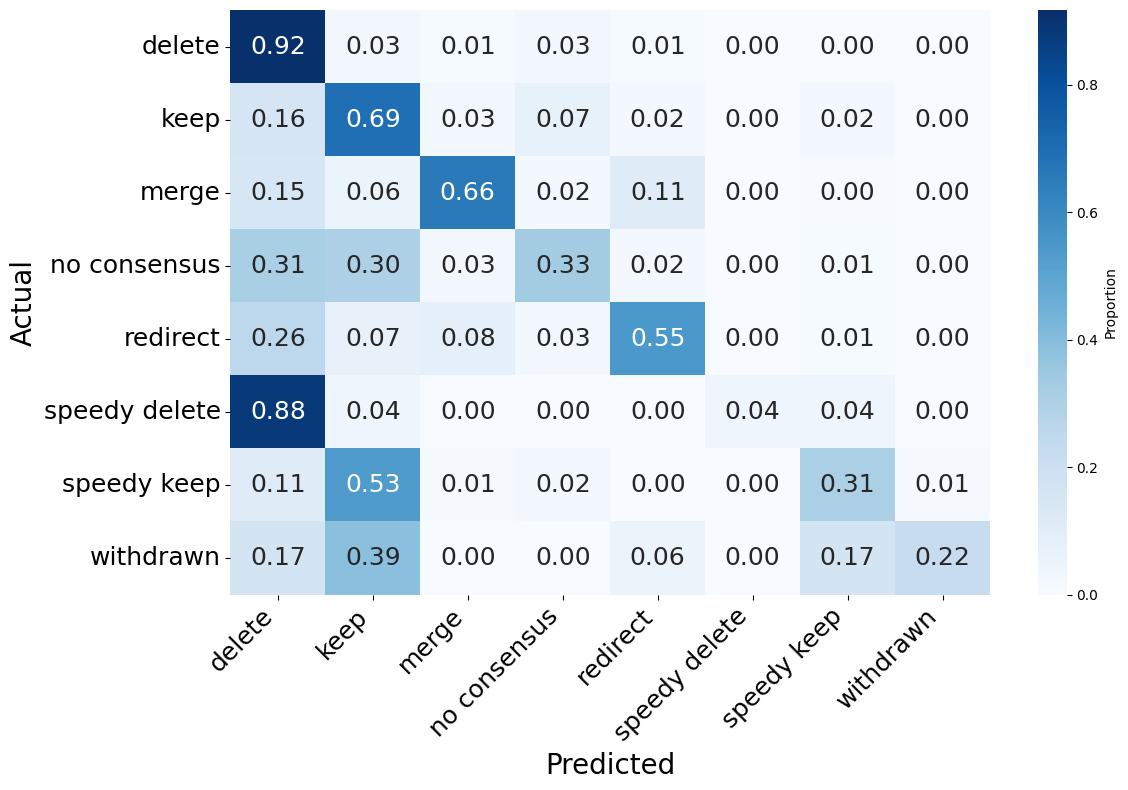} \\
        \text{(a) Fulltext Setting} \\
        \\
        \includegraphics[width=\columnwidth]{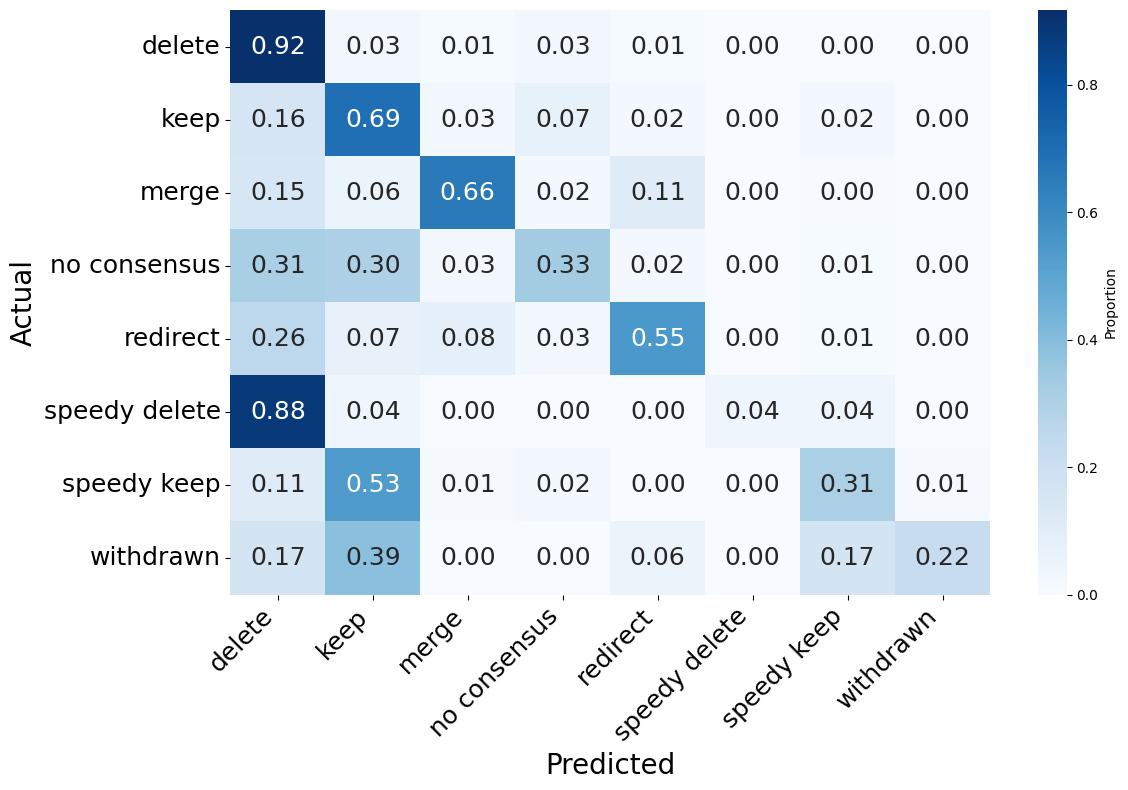} \\
        \text{(b) Masked Setting} \\
    \end{tabular}
    \caption{Confusion Matrix for RoBERTa-Large model in Outcome Prediction Task.}
    \label{fig:outcome_roberta}
\end{figure}

\subsubsection{In-platform}
\label{ref:inplatform}

We train each model with platform-specific training sets under both masked and full text settings, and report F1 results in Table \ref{tab:res_combined}. As expected, we can see that hiding the self-reported tags generally causes a drop in performance across the board, most noticeable in the X (Twitter)-specific model. We can also see that RoBERTa large is always the best model in full text, while this is more inconsistent in the masked setup. Further per-label analysis of the difference between the full text and masked settings for RoBERTa large (the best performing model) is provided in Figure \ref{fig:outcome_roberta}, which shows confusion matrices for the Wikipedia-en dataset. The performance drop for `keep', `merge' and `withdrawn' suggests that editors are more decisive about deletion of the article than keeping it. It also shows that full text is almost always useful, but interestingly, \texttt{merge} decisions benefit less from seeing these tags, likely because merging discussions are often less explicit and drift more between deletion and keep. Another interesting finding (which is consistent across both platforms) is that the \texttt{withdrawn} outcome often gets confused with \texttt{keep}, again reinforcing this idea of more ambiguity when the decision is \textit{not} leading towards deletion. Next, \texttt{no consensus} outputs seem very hard to predict, with an almost even split between predictions spread among the correct class (\texttt{no consensus}), \texttt{delete} and \texttt{keep}. And finally, a striking result is the massive confusion between \texttt{speedy delete}  and \texttt{delete} in the masked setting. This suggests that, in practice, there virtually no difference in how editors \textit{talk about} deleting articles, but they are however implicitly opinionated about how urgently the decisions needs to happen.

Another interesting perspective on this experiment is the option to explore more efficient approaches than simply fine-tuning an arguably large model, especially given that the size of the datasets is very varied. To test our hypothesis that simpler approaches could be beneficial, we evaluate a SetFit (Sentence-transformer Fine-tuning ) model \cite{tunstall2022efficient}. SetFit is a simple yet powerful technique that fine-tunes a sentence transformers model\footnote{We used \texttt{BAAI/bge-base-en-v1.5 } \cite{bge_embedding}, a model roughly 3 times smaller than our best performing fine-tuned models.} by artificially sampling training pairs for a contrastive learning stage, and uses the fine-tuned embeddings as feature vectors for a logistic regression classifier.  Note that, in SetFit, in the embedding fine-tuning stage a large number of document pairs can be generated, specifically $K(K-1)/2$, where $K$ is the number of labeled examples (i.e., the original training set). Therefore, we subsample the original training sets into a smaller stratified training set of 100 labeled examples. We found a striking boost in performance with this model, especially for smaller datasets (like Wikinews), which suggestst that, in production environments, SetFit could be an efficient and highly performing option. We show a SetFit vs best model comparison in Table \ref{tab:best_lm_results}.




\begin{table}[!h]
\centering
\footnotesize
\resizebox{0.95\columnwidth}{!}{%
\begin{tabular}{@{}llll@{}}
\toprule
\textbf{Platform}             & \textbf{Setting} & \textbf{Model} & \textbf{F1} \\ \midrule
\multirow{3}{*}{Wikipedia}    & Fulltext         & RoBERTa-L      & 0.60              \\
                              & Masked           & RoBERTa-L      & 0.52              \\
                              & Fulltext        & SetFit      & \textbf{0.65}      \\       
                              & Masked  & SetFit     & 0.57 \\ \midrule
\multirow{3}{*}{Wikidata-ent} & Fulltext         & RoBERTa-L      & 0.63              \\
                              & Masked           & RoBERTa-L      & 0.61              \\
                              & Fulltext        & SetFit     & \textbf{0.88}             \\ 
                              & Masked  & SetFit     &  0.87 \\\midrule
\multirow{3}{*}{Wikidata-pr}  & Fulltext         & RoBERTa-L      & 0.62              \\
                              & Masked           & BERT-L         & 0.60              \\
                              & Fulltext        & SetFit      & 0.61              \\ 
                              & Masked  & SetFit     &  \textbf{0.70} \\\midrule
\multirow{3}{*}{Wikinews}     & Fulltext         & RoBERTa-L      & 0.47              \\
                              & Masked           & RoBERTa-L      & 0.42              \\
                              & Fulltext        & SetFit      & \textbf{0.57}              \\ 
                              & Masked  & SetFit     &  0.44 \\\midrule
\multirow{3}{*}{Wikiquote}    & Fulltext         & RoBERTa-L      & 0.76              \\
                              & Masked           & BERT-L         & 0.72              \\
                              & Fulltext        & SetFit      & \textbf{0.87}              \\ 
                              & Masked  & SetFit     &  0.44 \\ \bottomrule
\end{tabular}%
}
\caption{Best model vs. SetFit results.}
\label{tab:best_lm_results}
\end{table}

\subsubsection{Cross-platform}
\label{sec:crossplatform}

In previous training experiments, we observe that models struggle to perform well in smaller datasets, likely due to the lack of training data (like Wikinews, where the performance was lowest on average for all models, with some cases up to 30\% drop - as in the case of the Twitter-specialized model). This motivates us to explore the potential of a cross-platform training regime, under the hypothesis that many features in deletion discussions might be similar across Wiki-platforms. We therefore perform an experiment where, for each model and training set, we evaluate on the test set of all the Wiki-platforms. Due to the variation in label sets, we simplify this experiment and map all the labels of each dataset to only \texttt{keep} and \texttt{delete}, the two common labels in all the datasets. We test the models in the fulltext setting.

The expectation from the results listed in Table \ref{tab:cross_model} would be to have a bold diagonal, i.e., a model trained on dataset X would be expected to be the best on the test set for X. While this is primarily the case, we find a comparable performance in other datasets, indicating a subtle but prominent generalization of the models across the platforms. The outlier in this trend is Wikinews, where we see both Wikipedia and Wikidata Entity-derived models performing better than the in-domain model. This can be attributed to the size of Wikinews dataset, which may not be enough for the models learn platform specific patterns. In fact, for this case, training on the most general dataset (i.e., Wikipedia) yields the best performance, specifically a non-negligible 11\% increase in F-1. The performances of Wikidata-entity and property across all other platform is also quite similar, despite of the large difference in data instances between them, further signaling the similarity in contents between the two important components of the same platform. However, there is an important difference, it seems that Wikidata properties transfers well into Wikidata entities (with only a 2\% drop in F1), however this is not the case vice versa, as a Wikidata entities-trained model falls short by 12\% vs the in-domain model (trained on Wikidata properties). 

Amongst all the similar performances from the models, an obvious outlier is Wikiquote, which fails to perform well on the other datasets, and all of the other models also fail to perform on the Wikiquote dataset, clearly showing the distinctive nature of Wikiquote discussions. However, it should be noted that compared to other popular platforms like Wikipedia, Wikiquote has a smaller editor base\footnote{According to Wikiquote's official Wikipedia page, it only has 474 active editors as compared to Wikipedia's  126,324 (in other words, Wikiquote has only 0.004\% of the editors of Wikipedia).}, which could cause a lack of diversity, consistency and overall quality, such as unmoderated discussions or inconsistencies between outcomes. 







\begin{table}[]
\resizebox{\columnwidth}{!}{
\begin{tabular}{lrrrrr}
\hline
                    & Wikip.            & Wikid.-ent       & Wikid.-pr  & Wikin.      & Wikiq.            \\ \hline
Wikip.           & \textbf{0.76} & 0.63           & 0.67          & 0.55  & 0.14          \\
Wikid.-ent     & 0.43          & \textbf{0.89} & 0.87         & 0.33 & 0.63          \\
Wikid.-pr & 0.61          & 0.71          & \textbf{0.83} & 0.04   & 0.1            \\
Wikin.            & \textbf{0.44} & 0.42           & 0.39           & 0.35  & 0.04        \\
Wikiq.           & 0.07          & 0.04           & 0.05          & 0.01   & \textbf{0.94} \\ \hline
\end{tabular}
}
\caption{Results of F1 scores of model performance on different test sets (columns represent the data models were trained on and rows represent the data model is tested on.).}
\label{tab:cross_model}
\end{table}

\begin{table}[!t]
\centering
\scriptsize
\resizebox{\columnwidth}{!}{%
\begin{tabular}{llrr}
\hline
\textbf{Language} & \textbf{Model}                     & \textbf{F1 (FT)} & \textbf{F1 (M)} \\ \hline
\multirow{4}{*}{gr} 
 & XLM-R-Base                  & 0.47                  & 0.38                \\
 & XLM-R-Large                  & 0.59         & 0.49                \\
 & MBERT       & 0.59         & 0.40                 \\
 & Tw.-XLM-R & 0.44                  & 0.40                 \\
 & SetFit & \textbf{0.81} & \textbf{0.60} \\ \hline
\multirow{4}{*}{es} 
 & XLM-R-Base                   & 0.66                  & 0.47                \\
 & XLM-R-Large                  & 0.88         & \textbf{0.85}                \\
 & MBERT       & 0.70                   & 0.67                \\
 & Tw.-XLM-R & 0.56                  & 0.46                \\
 & SetFit & \textbf{0.90} & 0.61  \\ \hline
\end{tabular}%
}
\caption{F1 score for Fulltext (FT) and Masked (M) settings for multilingual models in Spanish (es) and Greek (gr).}
\label{tab:multiling_wikipedia}
\end{table}




\subsubsection{Multilingual}
\label{sec:multilingual}

For multilingual datasets (Wikipedia-es and Wikipedia-gr), we experiment with XLM-R \cite{conneau2019unsupervised} (Base and Large, XLM-R-Base and XLM-R-Large), Multilingual BERT (MBERT) \cite{devlin2018bert}, and Twitter-XLM-R (Tw.-XLM-R) \cite{barbieri2020tweeteval} (to explore if a model specialized on Twitter, another instance of user-generated content, could give advantages). We also introduced SetFit in this experiment, on top of a multilingual embedding model\footnote{Specifically, \texttt{paraphrase-multilingual-MiniLM-L12-v2}.}. Due to the large data difference between the languages we consider (c.f. Table \ref{tab:overall_data_stats}), as well as them having a different set of outcome labels, a cross language comparison is perhaps not appropriate, and therefore we discuss classification results on both languages separately.

\begin{table*}[!t]
\resizebox{\textwidth}{!}{%
\begin{tabular}{@{}p{8cm}p{12cm}r@{}}
\toprule
\textbf{Policy}                         & \textbf{Example}                                         & \multicolumn{1}{l}{\textbf{Instances}} \\ \midrule
Wikipedia:Notability                    & {[}WP:N{]} Fails and with only routine coverage.        & 232,422 (70.22\%)                           \\
Wikipedia:What Wikipedia is not         & If there are sources linking this to the attack.        & 34,559 (10.44\%)                                 \\
Wikipedia:No original research          & "WP:OR, . Mishmash of random trivia.                    & 13,583 (4.10\%)                            \\
Wikipedia:Verifiability                 & I'm leaning delete per as this is completely unverified.& 12,531 (3.78\%)                           \\
Wikipedia:Arguments to avoid in deletion discussions & Her own work is admittedly ordinary, even run-of-the-mill. & 8,105 (2.44\%) \\
Wikipedia:Biographies of living persons & Because of BLP requirements, it needs to be rewritten.  & 7,346 (2.21\%)                         \\
Wikipedia:Criteria for speedy deletion  & In regards to the above comment about speedy deletion.  & 5,833 (1.77\%)                            \\
Wikipedia:Articles for deletion         & What searches did you do to establish notability?       & 5,758 (1.75\%)                             \\
Wikipedia:Wikipedia is not a dictionary & Clearly a lexical entry and in violation of policies.   & 5,474 (1.65\%)                                  \\
Wikipedia:Deletion policy               & In this case, I don't see the point of a redirect.      & 5,332 (1.61\%)                                 \\ \bottomrule
\end{tabular}%
}
\caption{Top 10 Policies of policy prediction task dataset with examples and number of instances (with percentage of the complete dataset).}
\label{tab:policy_sample}
\end{table*}

\begin{figure}[!h]
    \centering
    \begin{tabular}{c}
        \includegraphics[width=\columnwidth]{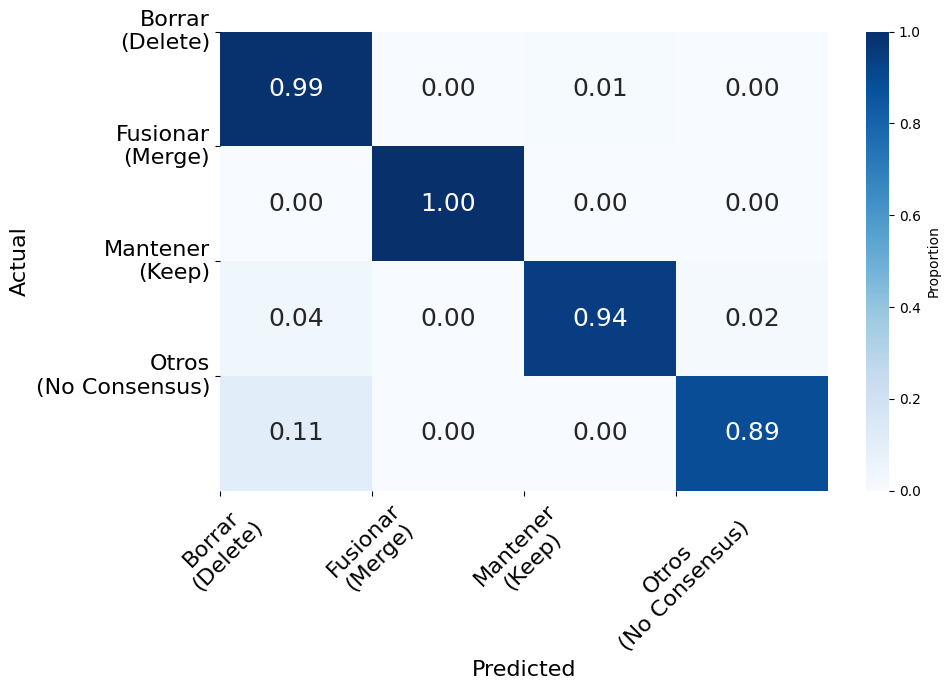 } \\
        \text{(a) Wikipedia-Es} \\
    \\
        
        \includegraphics[width=\columnwidth]{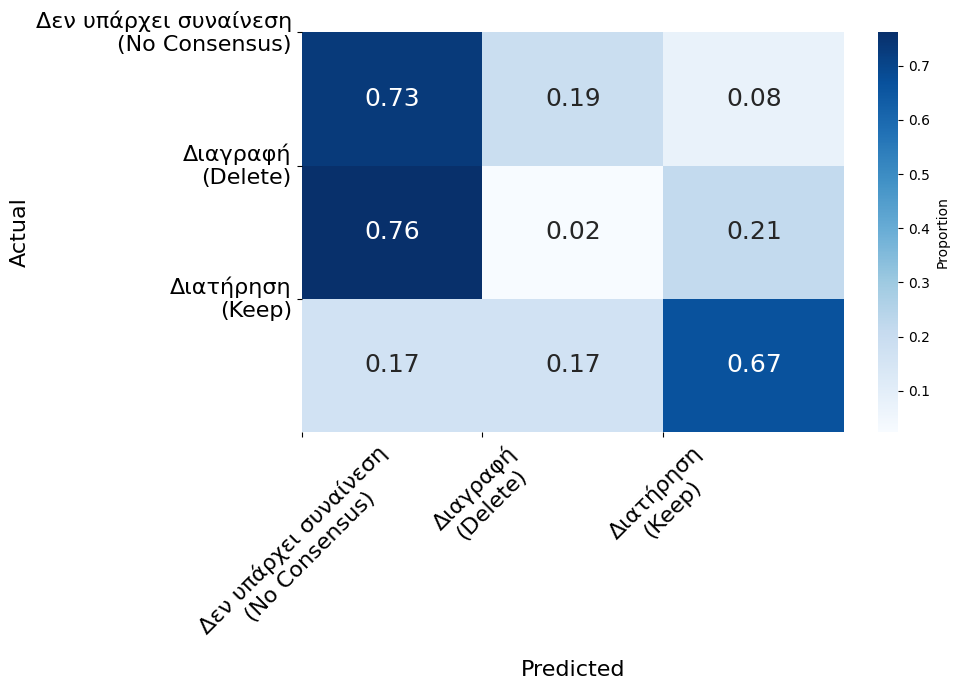 } \\
        \text{(b) Wikipedia-Gr} \\
    \end{tabular}
    \caption{Confusion Matrix for XLM-RoBERTa-Large model in Outcome Prediction Task for Spanish and Greek Wikipedia.}
    \label{fig:outcome_roberta_ml}
\end{figure}

Table \ref{tab:multiling_wikipedia} shows results following a similar performance pattern as the English experiment, with a significant difference between fulltext and masked setups. However, it is worth noting that XLM-R-Large and MBERT trained and tested on masked data in Spanish were able to come very close to the fulltext variant. This clearly points to Spanish editors using more explicit language when discussing whether an article should be deleted, giving a stronger signal to a classifier even after masking these self-assigned labels. This can be further verified in confusion matrices (on the fulltext setting, Figure \ref{fig:outcome_roberta_ml}), where the Greek model struggles to distinguish between \texttt{delete} and \texttt{no consensus}, which is certainly not the case in Spanish. Concerning the SetFit results, these are, again, surprisingly good, being the best model on the same test sets over fully fine-tuned models, with the exception only of the masked experiment, where it is outperformed by XLM-R-Large and MBERT. We attribute this to a potential mismatch of the style/topics/theme of the subsampled dataset and the training set. We leave for future work performing multiple runs to evalute the robustness of SetFit (or other approaches based on synthetic data generation) when datasets are varied.

\begin{figure}[!b]
    \centering
    \includegraphics[width=\columnwidth]{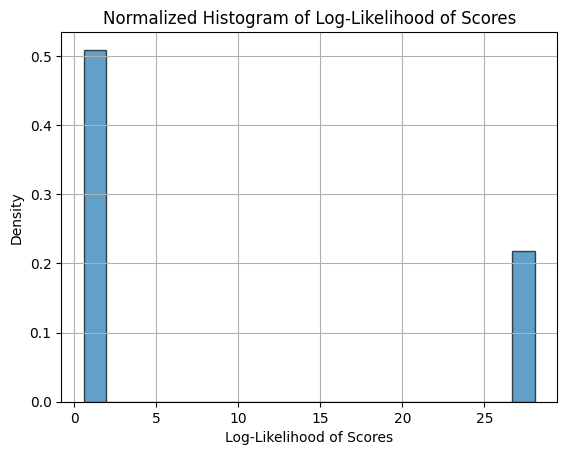}
    \caption{Normalized perplexity distribution for Twitter-XLM-Roberta in a sample of the Greek Wikipedia.}
    \label{fig:es_gr_ppl}
\end{figure}

Finally, we asked ourselves the question of why a strong multilingual model (XLM-R) further specialized on multilingual data from social media (Tw.-XLM-R) model would perform so poorly on another instance of user-generated texts which, as we can see from the examples in Tables \ref{tab:dataset_examples_all} and \ref{tab:policy_sample}, are not so different from well-formed tweets (note that in the original sampling from \citet{barbieri2020tweeteval} a few heuristics were put in place to filter out pure noise like all-emoji tweets). One approach to gain further insights is by computing pseudo (log) likelihood \cite{salazar2019masked} over sequences in the dataset from different models by taking a sample of the data, masking the actual (sub)word one at a time, and compute the loss of the models, and averaging over the whole sequence. Higher likelihood typically points to a model with a ``good grasp'' of the presented data (domain, style, themes, etc), as shown, e.g., in the context of temporal adaptation \cite{loureiro2022timelms}. To this end, we computed pseudo log likelihood over a sample of 1,000 Greek Wikipedia articles. We find the distribution in Figure \ref{fig:es_gr_ppl}, which, instead of a curve-shaped distribution (which would be the ideal), has two clear spikes, which suggests that this particular Twitter model might struggle to generalize (very low likelihood scores). Further analysis into the role of tokenization is left for future work.

\subsection{Stance and Policy prediction}

Stance detection classifies a moderator's opinion towards the article using stance labels (keep, delete, merge, comment), while policy prediction identifies an explicitly or implicitly mentioned Wikimedia policy (refer to Table \ref{tab:policy_sample} for illustrative examples). Both are comment level tasks. We experiment with the same base models from the Outcome Prediction task. We report our results in Table \ref{tab:stance-policy}, which shows results for both tasks according to weighted F1-score. We are interested primarily in this metric to understand the benefit of such models for the platform as a whole, rather than investigating nuances in individual categories which account, in practice, for a very small proportion of the dataset. We still perform per-label analysis, but we believe weighted F1 in this scenario sends a clearer takeaway message to those interested in automating content moderation in these platforms in production environments.

\subsubsection{Stance Detection}
\label{sec:stancedetection}

Similar to \citet{kaffee2023should} we pose our stance detection task as a 4-class classification with the labels \texttt{delete}, \texttt{keep}, \texttt{merge} and \texttt{comment}, where the first three labels carry the same meaning as the outcome prediction task, and \texttt{comment} means the discussion goes on. Our stance detection results are comparable to the ones reported in \citet{kaffee2023should}, which in fact points to their strong and robust model, since their reported results were in macro F1, and are only slightly lower than ours (80\% macro F1, vs our 83\% weighted F1). In terms of analysis, we do not find any major differences between RoBERTa-Large and BERT-Large, although as is the norm in this paper, Tw.-RoBERTa-B does not perform well. Following from our previous experiments, we also test the ability of SetFit in these tasks. In this case, we also downsample the training and validation sets, in this case, from the original to 1,000 (train) and 300 (validation) stratified samples, with the test set staying the same for enabling a comparison. While not clearly outperforming the other models, as it was the case in previous experiments, it turned out to be an extremely competitive option, rivaling fully fine-tuned PLMs.

\subsubsection{Policy Prediction}
\label{sec:policyprediction}

\begin{table}[!t]
\Large
\centering
\resizebox{\columnwidth}{!}{
\begin{tabular}{lrrrrrrrr}
\hline
                     & \multicolumn{4}{c}{\textbf{Stance}}                                                                                                           & \multicolumn{4}{c}{\textbf{Policy}}                                                                                                          \\ \cline{2-9} 
                     & \multicolumn{1}{c}{\textbf{Acc,}} & \multicolumn{1}{c}{\textbf{Prec.}} & \multicolumn{1}{c}{\textbf{Rec.}} & \multicolumn{1}{c|}{\textbf{F1}} & \multicolumn{1}{c}{\textbf{Acc,}} & \multicolumn{1}{c}{\textbf{Prec.}} & \multicolumn{1}{c}{\textbf{Rec.}} & \multicolumn{1}{c}{\textbf{F1}} \\ \hline
RoBERTa-B         & 0.90                               & 0.83                                 & 0.78                               & \multicolumn{1}{r|}{0.81}          & 0.83                                & 0.70                                 &0.56                                & 0.61                              \\
RoBERTa-L        & 0.94                                & 0.85                                 & 0.81                                & \multicolumn{1}{r|}{\textbf{0.83}}          & 0.86                                & 0.74                                 & 0.62                                & \textbf{0.67}                              \\
BERT-B            & 0.89                                & 0.81                                 & 0.80                                & \multicolumn{1}{r|}{0.80}          & 0.81                                & 0.65                                 & 0.49                                & 0.55                              \\
BERT-L           & 0.91                                & 0.84                                 & 0.82                                & \multicolumn{1}{r|}{\textbf{0.83}}          & 0.84                                & 0.71                                 & 0.59                                & 0.63                              \\
DistilBERT-B           & 0.90                                & 0.83                                 & 0.74                                & \multicolumn{1}{r|}{0.78}          & 0.80                                & 0.69                                 & 0.46           & 0.50                              \\
Tw.-RoBERTa-B & 0.88                                & 0.80                                & 0.66                                & \multicolumn{1}{r|}{0.70}          & 0.81                                & 0.68                                 & 0.53                                & 0.58                              \\ 
SetFit & 0.83 & 0.81 & 0.82 &  \multicolumn{1}{r|}{0.82} & 0.66 & 0.68 & 0.67 & \textbf{0.67}\\ \hline

\end{tabular}
}
\caption{Stance and policy prediction results (with Weighted-F1 scores).}
\label{tab:stance-policy}
\end{table}

\begin{figure}[!t]
    \centering
    \includegraphics[width=\columnwidth]{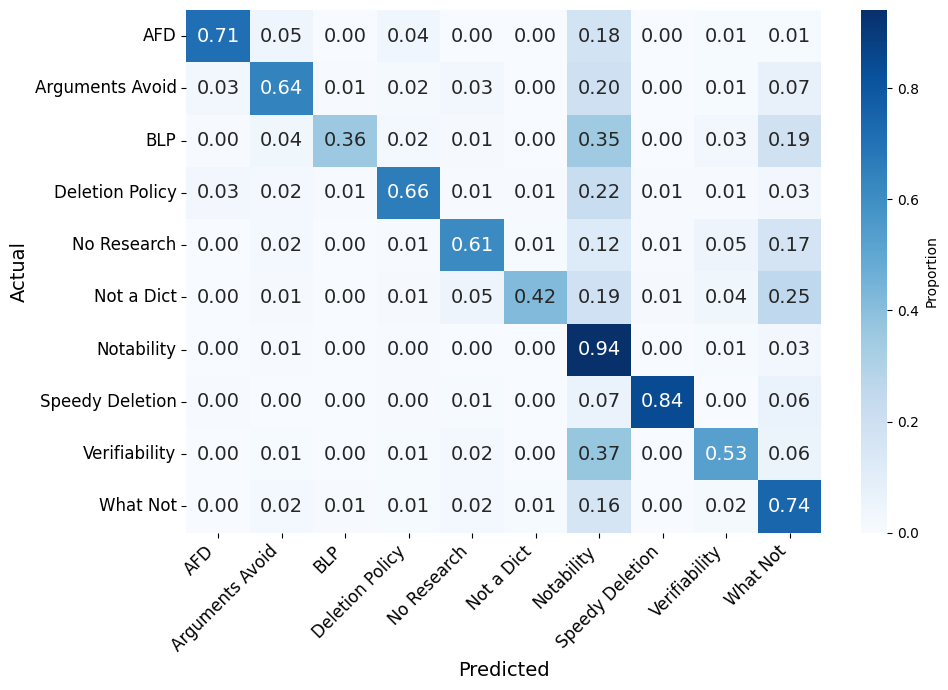}
    \caption{Confusion Matrix for RoBERTa-Large in Policy Prediction.}
    \label{fig:comfusion_policy}
\end{figure}

We modify the policy prediction task into a 10-label setup and follow a similar experimental setup as in previous sections. Our results show that with this task formulation works quite well, with 0.67 F1 for the best model (RoBERTa-Large) as shown in Table \ref{tab:stance-policy}. In terms of comparison with previous works, \citet{kaffee2023should} reported Accuracy figures about 0.75 on the original dataset (90+ labels), whereas we achieve about 10 points more in a trimmed down version. This suggests that the very long tail of about 80 infrequent labels and the likely under-performance on them does not impact the overall picture, and we can conclude that both their models and ours would behave similarly if deployed. The most interesting part of this experiment, however, is again looking at sources of confusion in the test set. In Figure \ref{fig:comfusion_policy} we see, for RoBERTa-Large, a well formed diagonal showing the correlation between actual and predicted labels. However, two major confounding sources emerge: `Notability' and `What Not' (shortened for `What Wikipedia is Not'). Notability is less surprising as this is the most frequent category, but `What Not' seems to be an overly generic category acting as a superset of other finer grained policies like `Not a Dict' (Wikipedia is Not a Dictionary). It would be interesting to explore the actual differences in comments pointing to these arguably interrelated policies, which could perhaps lead to merging them or further splitting `What Not' into others.

\section{Conclusion}

Automated Content moderation is a challenging yet important part of maintaining healthy content in community driven Wiki-platforms. Through this work, we analyze four different Wiki-platforms and three languages to give an all-round understanding of automated content moderation scenarios in these Wikis. Our analysis shows that these community based platforms can highly benefit from the usage of PLM based content moderation strategies, and to that end, we contribute a dataset and a range of strong baseline results from different PLMs for the community to build on.


\section*{Limitations}
\label{sec:limitations}
Our work does not extensively explore all deletion discussions obtainable from Wikipedia (throughout the years), even though it can be obtained using our package. We also do not explore any other LMs except BERT-family of models, and due to lack of domain data for sentiment analysis and offensive language detection, we do not train our own models for those tasks. Finally, the tool we propose here can be made better with integration of more analytical tasks and capabilities of model based activities, such as fine-tuning.

\section*{Ethics statement}
\label{sec:ethics}
We believe that enhancing quality control for Wikipedia and other sibling Wiki platforms, which is the most popular online encyclopedia through content moderation is always of utmost importance. There is importance of Wikipedia as a viable knowledge source for users, and a data source for today's NLP research is undeniable. This calls for the necessity of tools that enable automated content moderation, so that the discussions that happens behind the curtain of Wikipedia articles regarding its reliability should maintain its standard, while providing resolution for the disputed ones. 

\bibliography{custom}

\appendix

\section{Train/Validation/Test Splits}
\label{sec:app-data}

The data splits with number of instances used in this paper are described in Table \ref{tab:data_stats_splits}.

\begin{table}[!t]
\centering
\scriptsize
\renewcommand{\arraystretch}{1.2} 
\footnotesize 
\setlength{\tabcolsep}{3pt} 
\resizebox{0.95\columnwidth}{!}{%
\begin{tabular}{@{}l l l r r@{}}
\toprule
\textbf{Lang.}  & \textbf{Platform}     & \textbf{Data} & \multicolumn{1}{l}{\textbf{Rows}} & \multicolumn{1}{l}{\textbf{Total}}  \\ 
\midrule
\multirow{15}{*}{en} 
    & \multirow{3}{*}{Wikipedia}    
        & Train & \multicolumn{1}{r}{12,963}  & \multirow{3}{*}{18,528}  \\
    &   & Val   & \multicolumn{1}{r}{1,856}   &                         \\
    &   & Test  & \multicolumn{1}{r}{3,709}   &                         \\ 
    \cmidrule(l){2-5} 
    & \multirow{3}{*}{Wikidata-ent} 
        & Train & \multicolumn{1}{r}{248,871} & \multirow{3}{*}{355,428} \\
    &   & Val   & \multicolumn{1}{r}{35,558}  &                         \\
    &   & Test  & \multicolumn{1}{r}{70,999}  &                         \\ 
    \cmidrule(l){2-5} 
    & \multirow{3}{*}{Wikidata-pr}  
        & Train & \multicolumn{1}{r}{349}     & \multirow{3}{*}{498}    \\
    &   & Val   & \multicolumn{1}{r}{52}      &                         \\
    &   & Test  & \multicolumn{1}{r}{97}      &                         \\ 
    \cmidrule(l){2-5} 
    & \multirow{3}{*}{Wikinews}     
        & Train & \multicolumn{1}{r}{63}      & \multirow{3}{*}{91}     \\
    &   & Val   & \multicolumn{1}{r}{9}       &                         \\
    &   & Test  & \multicolumn{1}{r}{19}      &                         \\ 
    \cmidrule(l){2-5} 
    & \multirow{3}{*}{Wikiquote}    
        & Train & \multicolumn{1}{r}{484}     & \multirow{3}{*}{695}    \\
    &   & Val   & \multicolumn{1}{r}{69}      &                         \\
    &   & Test  & \multicolumn{1}{r}{142}     &                         \\ 
\midrule
\midrule
\multirow{3}{*}{es}  
    & \multirow{3}{*}{Wikipedia}    
        & Train & \multicolumn{1}{r}{2,291}   & \multirow{3}{*}{3,274}   \\
    &   & Val   & \multicolumn{1}{r}{294}     &                         \\
    &   & Test  & \multicolumn{1}{r}{689}     &                         \\ 
\midrule

\multirow{3}{*}{gr}  
    & \multirow{3}{*}{Wikipedia}    
        & Train & \multicolumn{1}{r}{274}     & \multirow{3}{*}{392}    \\
    &   & Val   & \multicolumn{1}{r}{35}      &                         \\
    &   & Test  & \multicolumn{1}{r}{83}      &                         \\ 
\midrule
\midrule

\multirow{3}{*}{en}  
    & \multirow{3}{*}{Stance}    
        & Train & \multicolumn{1}{r}{372,033} & \multirow{3}{*}{437,770} \\
    &   & Val   & \multicolumn{1}{r}{21,961}  &                         \\
    &   & Test  & \multicolumn{1}{r}{43,776}  &                         \\ 
\midrule

\multirow{3}{*}{en}  
    & \multirow{3}{*}{Policy}    
        & Train & \multicolumn{1}{r}{274,867} & \multirow{3}{*}{341,337} \\
    &   & Val   & \multicolumn{1}{r}{30,540}  &                         \\
    &   & Test  & \multicolumn{1}{r}{35,930}  &                         \\  
\bottomrule
\end{tabular}%
}
\caption{Data distribution statistics, divided in 3 blocks: English Outcome Prediction (top), Multilingual Outcome Prediction (middle), and (English) Stance and Policy Prediction (bottom).}
\label{tab:data_stats_splits}
\end{table}

\section{Training details and results for Outcome Prediction}
\label{app:finetuning_lm}

Following are the details of different hyperparameters we use in our experiments, which as can be seen vary across datasets due mostly to dataset size.

\begin{itemize}
    \item Number of epochs: 20 (Outcome Prediction)/ 5 (Stance/Policy detection)
    \item Learning rate: 1e-5 (Outcome Prediction)/ 2e-6 (Stance/Policy detection)
    \item Batch size: 4
    \item Optimizer: Adam
    \item Resource used: NVIDIA RTX 4090
\end{itemize}

\end{document}